\documentclass[10pt,conference]{IEEEtran}

\usepackage{amsmath,amssymb,amsfonts}
\usepackage{bm}
\usepackage{mathtools}
\usepackage{graphicx}
\usepackage{booktabs}
\usepackage{multirow}
\usepackage{xcolor}
\usepackage{algorithm}
\usepackage{algorithmic}
\usepackage{cite}
\usepackage{url}
\usepackage{hyperref}

\title{Empirical Minimal-Realisation Compression of Deep Neural Networks via Controllability--Observability Tests}

\author{\IEEEauthorblockN{Anis Hamadouche}
\IEEEauthorblockA{\textit{The School of Engineering \& Physical Sciences} \\
\textit{Heriot-Watt University}\\
Edinburgh EH14 4AS, UK \\
anis.hamadouche@hw.ac.uk}

\and

\IEEEauthorblockN{Amir Hussain}
\IEEEauthorblockA{\textit{SDAIA-KFUPM Joint Research Centre for Artificial Intelligence} \\
\textit{King Fahd University of Petroleum and Minerals}\\
Dhahran, Saudi Arabia  \\
amir.hussain@kfupm.edu.sa}
}

\begin{document}

\maketitle


\begin{abstract}
Deep neural networks often contain substantial hidden-state redundancy, but most compression methods operate directly on weights, neurons, or quantised representations without explicitly characterising the dynamical role of internal states. This paper proposes a controllability--observability framework for empirical state-order reduction of deep neural networks. By viewing a trained network as a depth-indexed nonlinear dynamical system, we construct data-driven reachability, observability, and balanced Gramians from hidden-state snapshots and output Jacobians. The resulting A/B/C tests estimate layer-wise reachable, observable, and jointly reachable--observable ranks. These ranks are then used not only as diagnostic measures of hidden-state redundancy, but also as actual compressed layer widths for realised reduced networks. Experiments on MNIST and CIFAR-10 compare the proposed balanced realisation against projection-based reduction, unstructured pruning, structured pruning, low-rank SVD, dynamic INT8 quantisation, and linear baselines. On MNIST, a four-layer SiLU DNN is reduced from state order $1024$ to $277$, giving $72.95\%$ state compression and $73.48\%$ parameter compression, while maintaining $95.45\%$ accuracy compared with $96.60\%$ for the full model. On CIFAR-10, a larger SiLU DNN is reduced from state order $4608$ to $1339$, giving $70.94\%$ state compression and $83.09\%$ parameter compression, while preserving accuracy from $54.45\%$ to $54.44\%$ and reducing CUDA inference latency by approximately $3\times$. The results show that balanced reachable--observable ranks provide a principled empirical minimal-realisation criterion for designing compact neural architectures with little or no loss in accuracy.
\end{abstract}

\begin{IEEEkeywords}
Deep neural network compression, controllability, observability, balanced truncation, empirical Gramians, minimal realisation, state-order reduction, pruning, low-rank compression, knowledge distillation.
\end{IEEEkeywords}

\section{Introduction}
\label{sec:introduction}

Deep neural networks (DNNs) have achieved strong performance across classification, perception, and representation-learning tasks, but their high parameter counts and wide hidden layers often make them expensive to deploy on resource-constrained hardware. As a result, neural network compression has become an important research direction. Existing methods include pruning, quantisation, low-rank factorisation, and knowledge distillation~\cite{han2015deepcompression,hinton2015distilling,denton2014exploiting,jacob2018quantization,liu2017networkslimming}. These approaches can significantly reduce storage or arithmetic cost, but they usually operate directly on weights, neurons, channels, or numerical precision. They do not explicitly answer a state-space question: which hidden-state directions are actually excited by the data distribution and which of those directions are relevant to the output?

This paper proposes a control-theoretic view of neural compression. We interpret a trained feedforward DNN as a depth-indexed nonlinear state-space system,
\begin{equation}
h_{\ell+1}=f_{\ell}(h_{\ell}), 
\qquad 
z=g(h_L),
\end{equation}
where $h_{\ell}$ denotes the hidden state at layer $\ell$ and $z$ denotes the output logits. This dynamical-systems interpretation is closely related to prior views of residual networks, neural ordinary differential equations, stable deep architectures, and optimal-control formulations of deep learning~\cite{he2016deep,haber2017stable,chen2018neural,ruthotto2020deep,benning2019deep}. However, instead of using this viewpoint primarily for stability or continuous-depth modelling, we use it to estimate the useful hidden-state order of a trained network.

The key idea is that not every hidden direction in a wide DNN is equally important. A direction may be excited by the input data but have little effect on the output logits. Conversely, a direction may be output-sensitive but rarely visited by the data distribution. Inspired by controllability, observability, and balanced truncation in model reduction~\cite{moore1981pca,antoulas2005approximation,lall2002subspace}, we propose three empirical tests for each hidden layer. Test A measures reachability using hidden-state snapshot covariances. Test B measures observability using output-logit Jacobians with respect to hidden states. Test C combines both effects through a balanced reachable--observable matrix. The resulting layer-wise C-balanced ranks estimate the number of hidden directions that are simultaneously reached by the data and observable at the output.

A central distinction in this work is between projected state reduction and realised neural compression. A projected reduced model maps hidden states into lower-dimensional coordinates using learned projection bases. This is useful for diagnosing the existence of a low-dimensional hidden representation, but it still stores the original full network. In contrast, the proposed realised C-balanced compression uses the C-balanced ranks directly as hidden-layer widths of a new compact DNN. Thus, a full network with widths $[n_1,n_2,\ldots,n_L]$ is replaced by a realised compressed architecture with widths $[r_1,r_2,\ldots,r_L]$, where $r_{\ell}\ll n_{\ell}$. This converts the empirical state-order estimate into a true parameter-compressed network.

The proposed framework differs from standard compression methods in both criterion and interpretation. Unstructured pruning removes individual weights but may not reduce dense inference latency. Structured pruning removes neurons or channels but often relies on local importance scores. Low-rank factorisation compresses weight matrices but does not explicitly preserve the nonlinear hidden-state geometry. Quantisation reduces numerical precision but leaves the hidden-state order unchanged. In contrast, the proposed method is state-based: it compresses the network according to directions that are jointly reachable from the data and observable through the classifier output.

We evaluate the proposed method on MNIST and CIFAR-10 using fully connected SiLU DNNs. On MNIST, the C-balanced test reduces the hidden-state order from $1024$ to $277$, corresponding to $72.95\%$ state compression. The realised C-balanced model reduces the parameter count from $400{,}906$ to $106{,}323$, giving $73.48\%$ parameter compression, while maintaining $95.45\%$ accuracy compared with $96.60\%$ for the full model. On CIFAR-10, the C-balanced test reduces the hidden-state order from $4608$ to $1339$, giving $70.94\%$ state compression. The realised compressed model reduces the parameter count from $9.97$M to $1.69$M, giving $83.09\%$ parameter compression, while preserving accuracy from $54.45\%$ to $54.44\%$ and reducing CUDA inference latency by approximately $3.0\times$.

The main contributions of this paper are summarised as follows:
\begin{itemize}
    \item We formulate feedforward DNNs as depth-indexed nonlinear state-space systems and define hidden-state order as a compression-relevant quantity distinct from parameter count.
    \item We propose empirical reachability, observability, and balanced reachable--observable tests for estimating layer-wise useful hidden-state dimensions.
    \item We introduce realised C-balanced neural compression, where C-balanced ranks are used as actual hidden-layer widths of compact networks.
    \item We compare the proposed method against projection-based reduction, unstructured pruning, structured pruning, low-rank SVD, dynamic INT8 quantisation, distillation, and linear baselines.
    \item We demonstrate substantial state-order and parameter reductions on MNIST and CIFAR-10 with little or no loss in accuracy.
\end{itemize}

The remainder of the paper is organised as follows. Section~\ref{sec:related_work} reviews neural compression, dynamical-systems views of DNNs, and empirical model reduction. Section~\ref{sec:dnn_state_space} formulates DNNs as nonlinear state-space systems and defines the compression objective. Section~\ref{sec:abc_tests} introduces the empirical A/B/C controllability--observability tests. Section~\ref{sec:realised_c_balanced_compression} presents the realised C-balanced compression procedure. Section~\ref{sec:experiments} reports the MNIST and CIFAR-10 experiments. Section~\ref{sec:limitations} discusses limitations and future extensions, and Section~\ref{sec:conclusion} concludes the paper.

\section{Related Work}
\label{sec:related_work}

This work connects neural network compression with control-theoretic model reduction. Existing compression methods usually reduce weights, neurons, channels, or numerical precision. In contrast, the proposed approach estimates the useful hidden-state order of a trained network using empirical reachability and observability tests. This section reviews the most relevant literature and positions the proposed method relative to prior work.

\subsection{Neural Network Compression}
\label{subsec:related_nn_compression}

Neural network compression has been widely studied as a way to reduce memory footprint, computational cost, and inference latency. A major class of methods is pruning, where redundant weights, neurons, filters, or channels are removed from a trained model. Deep Compression combines pruning, trained quantisation, and entropy coding to significantly reduce the storage requirements of neural networks~\cite{han2015deepcompression}. Unstructured pruning removes individual weights and can achieve high sparsity, but it does not necessarily reduce dense matrix dimensions or hardware latency unless sparse kernels are used. Structured pruning removes complete neurons, filters, or channels and is therefore more suitable for practical acceleration.

Channel-level pruning methods aim to produce thinner deployable networks. Network Slimming uses sparsity regularisation on batch-normalisation scale parameters to identify and remove insignificant channels~\cite{liu2017networkslimming}. Other pruning studies have also shown that the final compressed architecture can be more important than inheriting the original pruned weights. In particular, Liu \emph{et al.} argued that, for some structured pruning settings, training the pruned architecture from scratch can match or outperform fine-tuning inherited weights~\cite{liu2019rethinking}. This observation is closely related to our realised compression strategy: the C-balanced ranks are used to define the compressed architecture, which is then trained as a new compact model.

Low-rank factorisation provides another important compression direction. Denton \emph{et al.} exploited approximate linear structure in convolutional networks by replacing large linear or convolutional operations with low-rank approximations~\cite{denton2014exploiting}. Such methods reduce computation by factorising weight matrices or convolutional tensors. However, low-rank compression is usually layer-local and weight-centric. It does not explicitly account for whether the retained directions are excited by the input distribution or observable at the output. Our experiments show that naive layer-wise SVD can substantially degrade accuracy when compared with the proposed state-based C-balanced realisation.

Quantisation reduces the numerical precision of weights and activations. Integer-only quantisation schemes have been developed to enable efficient inference on low-power and mobile hardware~\cite{jacob2018quantization}. Quantisation can reduce memory bandwidth and improve deployment efficiency, but it leaves the hidden-state order and architecture width unchanged. Therefore, quantisation is complementary to the proposed method: C-balanced compression reduces the number of hidden coordinates, while quantisation reduces the numerical precision of the remaining computations.

Knowledge distillation compresses a large teacher model into a smaller student model by matching softened output distributions~\cite{hinton2015distilling}. Distillation is often effective because teacher logits contain additional class-similarity information beyond hard labels. In this paper, distillation is used as an optional training strategy for the realised C-balanced network. However, the main novelty is not the distillation loss itself, but the way the student architecture is chosen: its layer widths are determined by empirical balanced reachable--observable ranks.

\subsection{Dynamical-Systems Views of Deep Networks}
\label{subsec:related_dynamical_dnn}

A growing body of work interprets deep networks as dynamical systems evolving along the depth dimension. Residual networks can be written as
\begin{equation}
h_{\ell+1}=h_{\ell}+\phi_{\ell}(h_{\ell}),
\end{equation}
which resembles a forward-Euler discretisation of a continuous-time dynamical system~\cite{he2016deep}. This interpretation motivated stable architectures for deep neural networks, where forward propagation is analysed using tools from ordinary differential equations and numerical stability~\cite{haber2017stable}. Neural ordinary differential equations make this connection explicit by replacing a finite sequence of layers with a learned continuous-depth ODE~\cite{chen2018neural}. PDE-inspired interpretations have also been used to motivate new convolutional and residual architectures~\cite{ruthotto2020deep}. In addition, optimal-control formulations of deep learning interpret training as a control problem over a depth-indexed dynamical system~\cite{benning2019deep}.

These works mainly focus on stability, trainability, expressivity, or continuous-depth modelling. Our work uses the dynamical-systems viewpoint for a different purpose: estimating the useful hidden-state order of a trained network. By treating hidden activations as states, we can ask whether each hidden direction is reachable from the data distribution and observable through the output logits. This leads naturally to empirical controllability and observability tests.

\subsection{Classical Model Reduction and Balanced Truncation}
\label{subsec:related_model_reduction}

Model reduction is a classical topic in control theory and numerical linear algebra. The goal is to approximate a high-dimensional dynamical system by a lower-dimensional system that preserves the important input--output behaviour. Balanced truncation is a particularly influential method because it uses both controllability and observability Gramians to identify states that are simultaneously difficult to control and difficult to observe~\cite{moore1981pca,antoulas2005approximation}. In linear systems, balanced coordinates reveal state directions ranked by Hankel singular values, and truncating weak balanced directions yields a reduced-order model with input--output relevance.

The philosophy of balanced truncation is closely aligned with the objective of neural compression. A hidden direction should not be retained merely because it exists in a wide layer; it should be retained if it is both reached by the input distribution and relevant to the output. However, standard balanced truncation is formulated for dynamical systems with explicit input and output maps, often in linear time-invariant form. Deep neural networks are nonlinear, depth-indexed, data-driven systems, and their ``inputs'' are samples from a data distribution rather than control signals in the classical sense. Therefore, the classical theory cannot be applied directly without adaptation.

\subsection{Empirical Gramians and Nonlinear Model Reduction}
\label{subsec:related_empirical_gramians}

For nonlinear systems, empirical Gramians provide a practical way to estimate controllability and observability properties from simulations or data. Lall, Marsden, and Glavaski introduced empirical model reduction for controlled nonlinear systems using data-driven approximations of balanced truncation~\cite{lall2002subspace}. These methods extend the balanced-reduction idea beyond purely linear systems by using sampled state responses and output sensitivities.

The proposed method is inspired by this empirical-Gramian viewpoint, but adapts it to trained neural networks. Instead of simulating a physical nonlinear control system, we collect hidden-state snapshots generated by a trained DNN on a representative dataset. Reachability is approximated by the covariance of hidden activations, while observability is approximated by output-logit Jacobians with respect to hidden states. The balanced matrix combines these two quantities and produces a layer-wise rank estimate. This gives a data-dependent estimate of the useful hidden-state order of the trained network.

\subsection{Positioning of This Work}
\label{subsec:related_positioning}

The proposed framework differs from existing compression methods in three main ways. First, it is state-based rather than purely weight-based. Pruning, SVD, and quantisation operate mainly on weights or numerical representations, whereas the proposed method analyses hidden-state directions. Second, it combines data excitation and output relevance. Reachability alone measures which directions are visited by the data, while observability alone measures which directions can affect the logits. The C-balanced test retains directions that satisfy both criteria. Third, the method converts the estimated state ranks into realised compressed architectures. Thus, the proposed C-balanced ranks are not merely diagnostic; they define the layer widths of compact networks.

In summary, prior work has extensively studied neural compression, dynamical interpretations of deep learning, and balanced model reduction. The contribution of this paper is to connect these ideas through an empirical controllability--observability framework for neural compression. The resulting method provides a principled way to estimate hidden-state redundancy and to design compact DNN architectures with reduced state order, fewer parameters, and competitive accuracy.

\section{DNNs as Nonlinear State-Space Systems}
\label{sec:dnn_state_space}

\subsection{Depth-Indexed Dynamics}
Deep neural networks can be interpreted as nonlinear dynamical systems evolving along the layer index. This viewpoint is well established for residual networks, neural ordinary differential equations, and stable architectures for deep learning, where network depth is interpreted as a discrete or continuous evolution variable~\cite{he2016deep,haber2017stable,chen2018neural,ruthotto2020deep,benning2019deep}. In this work, we adopt this state-space interpretation for empirical state-order analysis and neural compression.

Let $x\in\mathbb{R}^{d}$ denote an input sample and let $h_{\ell}\in\mathbb{R}^{n_{\ell}}$ denote the hidden state at layer $\ell$. The input is treated as the initial state,
\begin{equation}
h_0=x.
\label{eq:initial_state}
\end{equation}
The forward propagation of a feedforward DNN can then be written as the depth-indexed nonlinear system
\begin{equation}
h_{\ell+1}=f_{\ell}(h_{\ell}),
\qquad
\ell=0,\ldots,L-1,
\label{eq:dnn_depth_dynamics}
\end{equation}
where $f_{\ell}:\mathbb{R}^{n_{\ell}}\rightarrow\mathbb{R}^{n_{\ell+1}}$ is the nonlinear transition map implemented by layer $\ell$. For a standard fully connected DNN, this map is
\begin{equation}
f_{\ell}(h_{\ell})
=
\sigma(W_{\ell}h_{\ell}+b_{\ell}),
\label{eq:dense_layer_transition}
\end{equation}
where $W_{\ell}$ and $b_{\ell}$ are trainable weights and biases, and $\sigma(\cdot)$ is a pointwise activation function such as ReLU, GELU, or SiLU.

The network output is obtained from the final hidden state through an output map
\begin{equation}
z=g(h_L),
\label{eq:dnn_output_map}
\end{equation}
where $z\in\mathbb{R}^{C}$ denotes the output logits and $C$ is the number of classes. For a linear classifier head,
\begin{equation}
g(h_L)=W_Lh_L+b_L.
\label{eq:linear_classifier_head}
\end{equation}
Thus, the full DNN can be represented compactly as
\begin{equation}
h_0=x,\qquad
h_{\ell+1}=f_{\ell}(h_{\ell}),\qquad
z=g(h_L).
\label{eq:dnn_state_space_form}
\end{equation}

This representation separates the hidden-state trajectory from the parameterisation of the layer maps. It therefore allows us to ask a control-theoretic question: which hidden-state directions are actually excited by the input distribution, and which of those directions are observable through the output logits?

\subsection{Relation to Residual and Continuous-Depth Networks}
The state-space interpretation is especially direct for residual networks. A residual block can be written as
\begin{equation}
h_{\ell+1}
=
h_{\ell}
+
\phi_{\ell}(h_{\ell}),
\label{eq:resnet_update}
\end{equation}
where $\phi_{\ell}$ denotes the residual transformation. This resembles a forward-Euler discretisation of a continuous-time dynamical system,
\begin{equation}
\frac{dh(t)}{dt}
=
\phi(h(t),t),
\label{eq:continuous_depth_ode}
\end{equation}
with the layer index playing the role of a discretised time variable~\cite{haber2017stable,chen2018neural,ruthotto2020deep}. Neural ODEs make this analogy explicit by replacing the discrete sequence of hidden layers with a learned differential equation whose solution defines the network output~\cite{chen2018neural}. Similarly, PDE-inspired interpretations of CNNs and ResNets have been used to design parabolic and hyperbolic network architectures~\cite{ruthotto2020deep}.

Although our experiments focus primarily on fully connected DNNs for controlled state-order analysis, the proposed formulation applies more broadly. For CNNs and ResNets, one may define the state at stage $\ell$ as a pooled feature representation,
\begin{equation}
h_{\ell}
=
\mathrm{GAP}(F_{\ell}(x)),
\label{eq:cnn_stage_state}
\end{equation}
where $F_{\ell}(x)$ is the feature map produced by stage $\ell$ and $\mathrm{GAP}(\cdot)$ denotes global average pooling. In this case, the hidden-state dimension corresponds to the number of channels at each stage, and state-order reduction becomes channel-width reduction.

\subsection{Hidden-State Order}
In this paper, the hidden-state order of a DNN is defined as the total number of hidden coordinates propagated through the network. For a network with $L$ hidden layers of widths $n_1,\ldots,n_L$, the full hidden-state order is
\begin{equation}
n_{\mathrm{state}}
=
\sum_{\ell=1}^{L}n_{\ell}.
\label{eq:full_state_order}
\end{equation}
This quantity is distinct from the number of trainable parameters. For a fully connected DNN with input dimension $d$, hidden widths $[n_1,\ldots,n_L]$, and output dimension $C$, the parameter count is
\begin{equation}
N_{\theta}
=
dn_1+n_1
+
\sum_{\ell=2}^{L}
(n_{\ell-1}n_{\ell}+n_{\ell})
+
n_LC+C.
\label{eq:dnn_parameter_count}
\end{equation}
The state order measures the dimension of the internal representation, whereas $N_{\theta}$ measures the number of stored trainable coefficients. These two quantities are related but not identical. A network may contain redundant hidden-state directions even when its weight matrices are not obviously low rank.

The proposed framework therefore focuses on hidden-state redundancy rather than only weight redundancy. A hidden direction is considered useful only if it is both excited by the data distribution and influential at the output. This motivates the empirical reachability and observability tests introduced in the next section.

\subsection{Compression Objective}
Let the full network have hidden widths
\begin{equation}
[n_1,n_2,\ldots,n_L].
\end{equation}
The objective is to identify reduced hidden widths
\begin{equation}
[r_1,r_2,\ldots,r_L],
\qquad
r_{\ell}\ll n_{\ell},
\label{eq:reduced_widths}
\end{equation}
such that the reduced model preserves the input--output function of the full network while reducing hidden-state order, parameter count, and inference latency.

The reduced hidden-state order is
\begin{equation}
r_{\mathrm{state}}
=
\sum_{\ell=1}^{L}r_{\ell},
\label{eq:reduced_state_order}
\end{equation}
and the state-compression ratio is
\begin{equation}
\eta_{\mathrm{state}}
=
1-
\frac{r_{\mathrm{state}}}{n_{\mathrm{state}}}.
\label{eq:state_compression_ratio}
\end{equation}
For a realised compressed fully connected DNN with reduced widths $[r_1,\ldots,r_L]$, the parameter count becomes
\begin{equation}
N_{\theta}^{\mathrm{red}}
=
dr_1+r_1
+
\sum_{\ell=2}^{L}
(r_{\ell-1}r_{\ell}+r_{\ell})
+
r_LC+C.
\label{eq:reduced_parameter_count}
\end{equation}
The corresponding parameter-compression ratio is
\begin{equation}
\eta_{\theta}
=
1-
\frac{N_{\theta}^{\mathrm{red}}}{N_{\theta}}.
\label{eq:parameter_compression_ratio}
\end{equation}

The compression problem may therefore be written as
\begin{equation}
\min_{\{r_{\ell}\}_{\ell=1}^{L}}
\quad
\sum_{\ell=1}^{L} r_{\ell}
\label{eq:compression_objective}
\end{equation}
subject to
\begin{equation}
\mathcal{A}(F_{\mathrm{red}})
\geq
\mathcal{A}(F_{\mathrm{full}})-\delta,
\label{eq:accuracy_constraint}
\end{equation}
where $\mathcal{A}(\cdot)$ denotes task accuracy and $\delta$ is an acceptable accuracy degradation. Directly solving this constrained problem is generally difficult. Instead, we estimate the useful hidden-state order at each layer using empirical controllability--observability tests. The resulting ranks are then used either as projection dimensions or as actual hidden widths in a realised compressed architecture.

\subsection{Projected Versus Realised State Reduction}
The proposed framework distinguishes between projected state reduction and realised neural compression. In projected reduction, the full hidden state is mapped to a lower-dimensional coordinate
\begin{equation}
q_{\ell}
=
U_{\ell}^{T}(h_{\ell}-\mu_{\ell}),
\qquad
q_{\ell}\in\mathbb{R}^{r_{\ell}},
\label{eq:projected_state}
\end{equation}
where $U_{\ell}\in\mathbb{R}^{n_{\ell}\times r_{\ell}}$ is a learned basis and $\mu_{\ell}$ is the empirical hidden-state mean. This representation is useful for diagnosing whether a low-dimensional hidden-state representation exists. However, it still requires storing the original network and the projection bases.

In realised reduction, the estimated ranks $r_{\ell}$ are used directly as the layer widths of a new compressed network:
\begin{equation}
\tilde h_{\ell+1}
=
\tilde f_{\ell}(\tilde h_{\ell}),
\qquad
\tilde h_{\ell}\in\mathbb{R}^{r_{\ell}}.
\label{eq:realised_reduction}
\end{equation}
This produces a genuinely smaller neural network with fewer hidden-state coordinates and fewer trainable parameters. The central question addressed in this paper is whether controllability--observability ranks estimated from the full network can provide reliable reduced widths for such realised compressed architectures.






\section{Empirical Controllability--Observability Tests}
\label{sec:abc_tests}

This section introduces the proposed empirical A/B/C tests for estimating the useful hidden-state order of a trained DNN. The tests are inspired by controllability, observability, and balanced truncation in dynamical systems, but are adapted to depth-indexed neural network dynamics and computed directly from hidden-state snapshots and output Jacobians.

Let a trained DNN be written as
\begin{equation}
h_0=x,\qquad
h_{\ell+1}=f_{\ell}(h_{\ell}),
\qquad
z=g(h_L),
\label{eq:abc_dnn_dynamics}
\end{equation}
where $h_{\ell}\in\mathbb{R}^{n_{\ell}}$ is the hidden state at layer $\ell$ and $z\in\mathbb{R}^{C}$ denotes the output logits. Given a set of $N$ input samples $\{x_i\}_{i=1}^{N}$, the corresponding hidden states at layer $\ell$ are collected as
\begin{equation}
H_{\ell}
=
\begin{bmatrix}
h_{\ell}(x_1)^T\\
h_{\ell}(x_2)^T\\
\vdots\\
h_{\ell}(x_N)^T
\end{bmatrix}
\in\mathbb{R}^{N\times n_{\ell}}.
\label{eq:hidden_snapshot_matrix}
\end{equation}
The empirical hidden-state mean is
\begin{equation}
\mu_{\ell}
=
\frac{1}{N}
\sum_{i=1}^{N}h_{\ell}(x_i),
\label{eq:hidden_state_mean}
\end{equation}
and the centred snapshot matrix is
\begin{equation}
\bar H_{\ell}
=
H_{\ell}
-
\mathbf{1}\mu_{\ell}^{T},
\label{eq:centered_snapshot_matrix}
\end{equation}
where $\mathbf{1}\in\mathbb{R}^{N}$ is the all-ones vector.

\subsection{Test A: Empirical Reachability}
\label{subsec:test_a_reachability}

The first test measures which hidden-state directions are excited by the data distribution. We define the empirical reachability Gramian at layer $\ell$ as the hidden-state covariance
\begin{equation}
G_{c,\ell}
=
\frac{1}{N}
\bar H_{\ell}^{T}\bar H_{\ell}
\in\mathbb{R}^{n_{\ell}\times n_{\ell}}.
\label{eq:reachability_gramian}
\end{equation}
Large eigenvalues of $G_{c,\ell}$ correspond to directions in hidden-state space that are strongly excited by the input data. Let
\begin{equation}
G_{c,\ell}
=
V_{c,\ell}\Lambda_{c,\ell}V_{c,\ell}^{T},
\label{eq:reachability_eigendecomp}
\end{equation}
where the eigenvalues are sorted in descending order,
\begin{equation}
\lambda_{c,\ell,1}
\geq
\lambda_{c,\ell,2}
\geq
\cdots
\geq
\lambda_{c,\ell,n_{\ell}}
\geq 0.
\end{equation}
The empirical reachable rank $r_{c,\ell}$ is selected as the smallest integer satisfying
\begin{equation}
\frac{
\sum_{i=1}^{r_{c,\ell}}\lambda_{c,\ell,i}
}{
\sum_{i=1}^{n_{\ell}}\lambda_{c,\ell,i}
}
\geq
1-\varepsilon.
\label{eq:reachability_rank}
\end{equation}
The corresponding reachability basis is
\begin{equation}
U_{c,\ell}
=
V_{c,\ell}(:,1:r_{c,\ell}).
\label{eq:reachability_basis}
\end{equation}

Test A is therefore a data-excitation test. It identifies the low-dimensional subspace that captures most of the hidden-state variance induced by the input distribution. However, a direction may be strongly excited by the data while having little effect on the output logits. For this reason, reachability alone is not sufficient for task-aware compression.

\subsection{Test B: Empirical Observability}
\label{subsec:test_b_observability}

The second test measures which hidden-state directions influence the output logits. For each input sample $x_i$, we define the output Jacobian with respect to the hidden state at layer $\ell$:
\begin{equation}
J_{\ell}(x_i)
=
\frac{\partial z(x_i)}{\partial h_{\ell}(x_i)}
\in\mathbb{R}^{C\times n_{\ell}}.
\label{eq:output_jacobian}
\end{equation}
The empirical observability Gramian is then defined as
\begin{equation}
G_{o,\ell}
=
\frac{1}{N}
\sum_{i=1}^{N}
J_{\ell}(x_i)^{T}J_{\ell}(x_i)
\in\mathbb{R}^{n_{\ell}\times n_{\ell}}.
\label{eq:observability_gramian}
\end{equation}
For any direction $v\in\mathbb{R}^{n_{\ell}}$, the quadratic form
\begin{equation}
v^{T}G_{o,\ell}v
=
\frac{1}{N}
\sum_{i=1}^{N}
\|J_{\ell}(x_i)v\|_2^2
\label{eq:observability_quadratic}
\end{equation}
measures the average sensitivity of the output logits to perturbations in the hidden-state direction $v$. Thus, directions associated with large eigenvalues of $G_{o,\ell}$ are highly observable at the output.

Let
\begin{equation}
G_{o,\ell}
=
V_{o,\ell}\Lambda_{o,\ell}V_{o,\ell}^{T},
\label{eq:observability_eigendecomp}
\end{equation}
with eigenvalues sorted in descending order. The empirical observable rank $r_{o,\ell}$ is selected as the smallest integer satisfying
\begin{equation}
\frac{
\sum_{i=1}^{r_{o,\ell}}\lambda_{o,\ell,i}
}{
\sum_{i=1}^{n_{\ell}}\lambda_{o,\ell,i}
}
\geq
1-\varepsilon.
\label{eq:observability_rank}
\end{equation}
The corresponding observability basis is
\begin{equation}
U_{o,\ell}
=
V_{o,\ell}(:,1:r_{o,\ell}).
\label{eq:observability_basis}
\end{equation}

Test B is task-aware because it depends on the output map and the classifier logits. However, observability alone can retain directions that are highly sensitive at the output but are rarely excited by the actual data distribution. Therefore, a balanced criterion is needed to retain directions that are both data-reachable and output-observable.

\subsection{Test C: Balanced Reachable--Observable Rank}
\label{subsec:test_c_balanced}

The third test combines reachability and observability. For each layer $\ell$, we define the empirical balanced matrix
\begin{equation}
B_{\ell}
=
G_{c,\ell}^{1/2}
G_{o,\ell}
G_{c,\ell}^{1/2}
\in\mathbb{R}^{n_{\ell}\times n_{\ell}},
\label{eq:balanced_matrix}
\end{equation}
where $G_{c,\ell}^{1/2}$ denotes the symmetric positive semidefinite square root of $G_{c,\ell}$. For any direction $w\in\mathbb{R}^{n_{\ell}}$, the energy induced by $B_{\ell}$ reflects the output sensitivity of directions weighted by how strongly they are excited by the data distribution.

Let
\begin{equation}
B_{\ell}
=
V_{b,\ell}\Lambda_{b,\ell}V_{b,\ell}^{T},
\label{eq:balanced_eigendecomp}
\end{equation}
where the eigenvalues
\begin{equation}
\lambda_{b,\ell,1}
\geq
\lambda_{b,\ell,2}
\geq
\cdots
\geq
\lambda_{b,\ell,n_{\ell}}
\geq 0
\end{equation}
are sorted in descending order. The C-balanced rank $r_{b,\ell}$ is selected as the smallest integer satisfying
\begin{equation}
\frac{
\sum_{i=1}^{r_{b,\ell}}\lambda_{b,\ell,i}
}{
\sum_{i=1}^{n_{\ell}}\lambda_{b,\ell,i}
}
\geq
1-\varepsilon.
\label{eq:balanced_rank}
\end{equation}

A corresponding state-space basis can be formed as
\begin{equation}
U_{b,\ell}
=
\mathrm{orth}
\left(
G_{c,\ell}^{1/2}
V_{b,\ell}(:,1:r_{b,\ell})
\right),
\label{eq:balanced_basis}
\end{equation}
where $\mathrm{orth}(\cdot)$ denotes column orthonormalisation. The projected C-balanced state is then
\begin{equation}
q_{\ell}
=
U_{b,\ell}^{T}(h_{\ell}-\mu_{\ell}),
\qquad
q_{\ell}\in\mathbb{R}^{r_{b,\ell}}.
\label{eq:c_balanced_projected_state}
\end{equation}

The C-balanced test is the primary compression criterion used in this paper. Unlike Test A, it does not retain directions solely because they are excited by the input. Unlike Test B, it does not retain directions solely because they are locally output-sensitive. Instead, it keeps directions that are jointly reachable from the data distribution and observable through the classifier output.

\subsection{Rank Selection and State-Order Estimate}
\label{subsec:rank_selection_state_order}

For each test, the same spectral energy criterion is used to select the layer-wise rank. Given a positive semidefinite matrix $M_{\ell}$ with eigenvalues
\begin{equation}
\lambda_{\ell,1}
\geq
\lambda_{\ell,2}
\geq
\cdots
\geq
\lambda_{\ell,n_{\ell}}
\geq 0,
\end{equation}
the selected rank is
\begin{equation}
r_{\ell}
=
\min
\left\{
r:
\frac{
\sum_{i=1}^{r}\lambda_{\ell,i}
}{
\sum_{i=1}^{n_{\ell}}\lambda_{\ell,i}
}
\geq
1-\varepsilon
\right\}.
\label{eq:general_rank_selection}
\end{equation}
The estimated reduced hidden-state order is then
\begin{equation}
r_{\mathrm{state}}
=
\sum_{\ell=1}^{L}r_{\ell}.
\label{eq:estimated_reduced_order}
\end{equation}
The corresponding state-compression ratio is
\begin{equation}
\eta_{\mathrm{state}}
=
1-
\frac{
r_{\mathrm{state}}
}{
n_{\mathrm{state}}
}.
\label{eq:abc_state_compression}
\end{equation}

The tolerance $\varepsilon$ controls the trade-off between compression and fidelity. Smaller values of $\varepsilon$ retain more spectral energy and therefore produce larger ranks, while larger values produce more aggressive compression. In all reported experiments, we use $\varepsilon=10^{-4}$ unless otherwise stated.

\subsection{Interpretation of the A/B/C Tests}
\label{subsec:abc_interpretation}

The three tests have complementary interpretations. Test A measures data excitation. If $r_{c,\ell}$ is close to $n_{\ell}$, then the hidden states at layer $\ell$ occupy a high-dimensional subspace under the data distribution. Test B measures task sensitivity. If $r_{o,\ell}$ is small, then only a small number of hidden-state directions significantly affect the output logits. Test C measures the dimension of the jointly reachable--observable subspace. A small C-balanced rank indicates that the layer contains many directions that are either weakly excited, weakly observable, or both.

This distinction is important for neural compression. Weight magnitude or matrix rank alone does not determine whether a hidden direction is useful for the input--output task. A hidden direction should be preserved only if it is both activated by the data and relevant to the output. The C-balanced rank therefore provides an empirical minimal-realisation-style estimate of the useful hidden-state order of the trained network.

\subsection{Computational Considerations}
\label{subsec:abc_computation}

Computing the reachability Gramian requires only forward passes and hidden-state snapshots. Its cost is therefore dominated by the collection of hidden activations and the eigendecomposition of matrices of size $n_{\ell}\times n_{\ell}$. Computing the observability Gramian is more expensive because it requires output Jacobians with respect to hidden states. In the experiments, this is computed on a subset of samples to reduce cost.

For classification with $C$ output logits, the observability Gramian can be written as
\begin{equation}
G_{o,\ell}
=
\frac{1}{N}
\sum_{i=1}^{N}
\sum_{c=1}^{C}
\nabla_{h_{\ell}}z_c(x_i)
\nabla_{h_{\ell}}z_c(x_i)^{T}.
\label{eq:observability_class_sum}
\end{equation}
Since $C$ is small for MNIST and CIFAR-10, exact Jacobian-based observability is feasible. For larger output spaces, one may use stochastic trace estimators, sampled logits, random output projections, or minibatch approximations.

\subsection{Algorithm}
\label{subsec:abc_algorithm}

Algorithm~\ref{alg:abc_tests} summarises the empirical A/B/C rank estimation procedure. The output of the algorithm is a set of layer-wise ranks for reachability, observability, and balanced compression. In the following section, the C-balanced ranks are used to construct realised compressed neural networks.

\begin{algorithm}[!t]
\caption{Empirical A/B/C Controllability--Observability Tests}
\label{alg:abc_tests}
\begin{algorithmic}[1]
\REQUIRE Trained network $F$, dataset $\mathcal{D}$, tolerance $\varepsilon$
\ENSURE Layer-wise ranks $\{r_{c,\ell},r_{o,\ell},r_{b,\ell}\}_{\ell=1}^{L}$
\FOR{$\ell=1$ to $L$}
    \STATE Collect hidden states $\{h_{\ell}(x_i)\}_{i=1}^{N}$
    \STATE Form centred snapshot matrix $\bar H_{\ell}$
    \STATE Compute reachability Gramian
    \[
    G_{c,\ell}=\frac{1}{N}\bar H_{\ell}^{T}\bar H_{\ell}
    \]
    \STATE Compute output Jacobians
    \[
    J_{\ell}(x_i)=\frac{\partial z(x_i)}{\partial h_{\ell}(x_i)}
    \]
    \STATE Compute observability Gramian
    \[
    G_{o,\ell}
    =
    \frac{1}{N}
    \sum_{i=1}^{N}
    J_{\ell}(x_i)^TJ_{\ell}(x_i)
    \]
    \STATE Form balanced matrix
    \[
    B_{\ell}
    =
    G_{c,\ell}^{1/2}G_{o,\ell}G_{c,\ell}^{1/2}
    \]
    \STATE Select $r_{c,\ell}$ from eigenvalues of $G_{c,\ell}$
    \STATE Select $r_{o,\ell}$ from eigenvalues of $G_{o,\ell}$
    \STATE Select $r_{b,\ell}$ from eigenvalues of $B_{\ell}$
\ENDFOR
\RETURN $\{r_{c,\ell},r_{o,\ell},r_{b,\ell}\}_{\ell=1}^{L}$
\end{algorithmic}
\end{algorithm}

\section{Realised C-Balanced Neural Compression}
\label{sec:realised_c_balanced_compression}

The A/B/C tests described in the previous section provide empirical estimates of the useful hidden-state dimension at each layer. In particular, the C-balanced test identifies directions that are both reachable from the data distribution and observable through the output logits. This section describes how these C-balanced ranks are converted into actual compressed neural architectures.

\subsection{Projected C-Balanced Reduction}
\label{subsec:projected_c_balanced}

Let $U_{\ell}\in\mathbb{R}^{n_{\ell}\times r_{\ell}}$ denote the C-balanced projection basis at layer $\ell$, where $r_{\ell}$ is the selected C-balanced rank. A projected reduced hidden coordinate is defined as
\begin{equation}
q_{\ell}
=
U_{\ell}^{T}(h_{\ell}-\mu_{\ell}),
\qquad
q_{\ell}\in\mathbb{R}^{r_{\ell}},
\label{eq:projected_c_state}
\end{equation}
where $\mu_{\ell}$ is the empirical mean of the hidden states at layer $\ell$. The corresponding approximate reconstruction is
\begin{equation}
\hat h_{\ell}
=
\mu_{\ell}+U_{\ell}q_{\ell}.
\label{eq:projected_c_reconstruction}
\end{equation}

The projected model is useful for testing whether the full network admits a low-dimensional hidden representation. However, it is not a true parameter-compressed model, because it still requires the original full network parameters and the additional projection bases. Its storage cost is therefore approximately
\begin{equation}
N_{\mathrm{proj}}
=
N_{\theta}
+
\sum_{\ell=1}^{L}
(n_{\ell}r_{\ell}+n_{\ell}),
\label{eq:projected_storage}
\end{equation}
where $N_{\theta}$ is the parameter count of the full network. Thus, projected C-balanced reduction diagnoses state redundancy but does not by itself guarantee a smaller deployable model.

\subsection{Realised C-Balanced Architecture}
\label{subsec:realised_architecture}

To obtain a genuinely compressed neural network, we use the C-balanced ranks directly as hidden-layer widths. If the full network has hidden widths
\begin{equation}
[n_1,n_2,\ldots,n_L],
\end{equation}
the realised compressed network is defined with widths
\begin{equation}
[r_1,r_2,\ldots,r_L],
\qquad
r_{\ell}\ll n_{\ell}.
\label{eq:realised_c_widths}
\end{equation}
For a fully connected DNN, the realised compressed dynamics are
\begin{equation}
\tilde h_{\ell+1}
=
\sigma(\tilde W_{\ell}\tilde h_{\ell}+\tilde b_{\ell}),
\qquad
\ell=0,\ldots,L-1,
\label{eq:realised_c_dynamics}
\end{equation}
where $\tilde h_{\ell}\in\mathbb{R}^{r_{\ell}}$ for $\ell\geq 1$ and $\tilde h_0=x$. The output logits are
\begin{equation}
\tilde z
=
\tilde W_L\tilde h_L+\tilde b_L.
\label{eq:realised_c_logits}
\end{equation}

The realised C-balanced model has hidden-state order
\begin{equation}
r_{\mathrm{state}}
=
\sum_{\ell=1}^{L}r_{\ell},
\label{eq:realised_state_order}
\end{equation}
and, for a fully connected architecture, parameter count
\begin{equation}
N_{\theta}^{\mathrm{C}}
=
d r_1+r_1
+
\sum_{\ell=2}^{L}
(r_{\ell-1}r_{\ell}+r_{\ell})
+
r_LC+C,
\label{eq:realised_c_parameter_count}
\end{equation}
where $d$ is the input dimension and $C$ is the number of output classes. Unlike projected reduction, the realised C-balanced model no longer stores the full hidden layers or projection bases. Therefore, it provides a true reduction in both state order and parameter count.

\subsection{Training the Realised C-Balanced Model}
\label{subsec:training_realised_c}

The realised C-balanced model can be trained in two ways. The first option is training from scratch using the standard supervised cross-entropy loss
\begin{equation}
\mathcal{L}_{\mathrm{CE}}
=
-\frac{1}{N}
\sum_{i=1}^{N}
\log
\frac{
\exp(\tilde z_{i,y_i})
}{
\sum_{c=1}^{C}\exp(\tilde z_{i,c})
}.
\label{eq:realised_ce_loss}
\end{equation}
This tests whether the C-balanced ranks define a compact architecture that is sufficiently expressive on its own.

The second option is knowledge distillation from the full model. Let $z_T$ and $z_S$ denote the logits of the teacher and student networks, respectively. The distilled loss is
\begin{equation}
\mathcal{L}_{\mathrm{distill}}
=
\alpha \mathcal{L}_{\mathrm{CE}}
+
(1-\alpha)T^2
D_{\mathrm{KL}}
\left(
\mathrm{softmax}\left(\frac{z_T}{T}\right)
\middle\|
\mathrm{softmax}\left(\frac{z_S}{T}\right)
\right),
\label{eq:distillation_loss}
\end{equation}
where $T$ is the distillation temperature and $\alpha\in[0,1]$ controls the balance between hard labels and teacher soft targets. Distillation encourages the compressed network to reproduce the smoother class-probability structure learned by the full model.

\subsection{Algorithm}
\label{subsec:realised_algorithm}

The overall procedure is summarised in Algorithm~\ref{alg:realised_c_balanced}. First, a full network is trained. Then, hidden-state snapshots and output Jacobians are collected on a representative data subset. These are used to compute empirical reachability and observability Gramians. The C-balanced matrix is then formed for each layer, and the energy criterion is used to select the C-balanced rank. Finally, a realised compressed network is instantiated using these ranks as its hidden-layer widths and trained either from scratch or by distillation.

\begin{algorithm}[!t]
\caption{Realised C-Balanced Neural Compression}
\label{alg:realised_c_balanced}
\begin{algorithmic}[1]
\REQUIRE Trained full network $F$, dataset $\mathcal{D}$, rank tolerance $\varepsilon$
\ENSURE Realised compressed network $F_{\mathrm{C}}$
\STATE Collect hidden-state snapshots $\{h_{\ell}(x_i)\}_{i=1}^{N}$ for each layer $\ell$
\STATE Compute the empirical reachability Gramian
\[
G_{c,\ell}
=
\frac{1}{N}
\bar H_{\ell}^{T}\bar H_{\ell}
\]
\STATE Compute output Jacobians
\[
J_{\ell}(x_i)=\frac{\partial z(x_i)}{\partial h_{\ell}(x_i)}
\]
\STATE Compute the empirical observability Gramian
\[
G_{o,\ell}
=
\frac{1}{N}
\sum_{i=1}^{N}
J_{\ell}(x_i)^TJ_{\ell}(x_i)
\]
\STATE Form the balanced matrix
\[
B_{\ell}
=
G_{c,\ell}^{1/2}G_{o,\ell}G_{c,\ell}^{1/2}
\]
\STATE Select the smallest $r_{\ell}$ satisfying
\[
\frac{\sum_{i=1}^{r_{\ell}}\lambda_i}
{\sum_{i=1}^{n_{\ell}}\lambda_i}
\geq 1-\varepsilon
\]
\STATE Instantiate $F_{\mathrm{C}}$ with hidden widths $[r_1,\ldots,r_L]$
\STATE Train $F_{\mathrm{C}}$ from scratch or using distillation from $F$
\RETURN $F_{\mathrm{C}}$
\end{algorithmic}
\end{algorithm}

\subsection{Comparison with Standard Compression Methods}
\label{subsec:comparison_compression_methods}

The realised C-balanced approach differs from standard compression methods in its compression criterion. Unstructured pruning removes individual weights, typically according to magnitude, but it does not necessarily reduce dense layer dimensions or hardware latency. Structured pruning removes neurons or channels, but often requires a separate importance heuristic and fine-tuning stage. Low-rank SVD factorises weight matrices, but it is a layer-local operation and may not preserve the nonlinear input--output representation. Quantisation reduces numerical precision but leaves the hidden-state order unchanged.

In contrast, C-balanced compression is state-based rather than weight-based. It estimates which hidden directions are jointly excited by the data and relevant to the output. The resulting ranks define the layer widths of a compact architecture. Therefore, the method reduces the hidden-state order directly and can also reduce parameter count and latency when implemented as a realised network.

\subsection{Extension to CNNs and ResNets}
\label{subsec:cnn_resnet_extension}

Although the main experiments in this paper focus on fully connected DNNs for controlled state-order analysis, the same principle extends naturally to convolutional and residual networks. For CNNs and ResNets, the state at stage $\ell$ can be defined as a global-average-pooled feature vector
\begin{equation}
h_{\ell}
=
\mathrm{GAP}(F_{\ell}(x)),
\label{eq:cnn_gap_state}
\end{equation}
where $F_{\ell}(x)$ is the feature map produced by stage $\ell$. In this setting, the C-balanced rank $r_{\ell}$ estimates the number of useful channels at stage $\ell$. Thus, the realised C-balanced compression rule becomes a channel-width reduction rule:
\begin{equation}
[C_1,C_2,\ldots,C_L]
\rightarrow
[r_1,r_2,\ldots,r_L].
\label{eq:cnn_channel_compression}
\end{equation}
This provides a pathway for extending the proposed framework from dense hidden-state compression to channel compression in CNN and ResNet architectures.

\subsection{Interpretation}
\label{subsec:realised_interpretation}

The realised C-balanced model can be interpreted as an empirical minimal-realisation approximation of the trained network. The full network may contain many hidden directions that are reachable but weakly observable, or observable but rarely excited by the data distribution. The C-balanced criterion retains directions that are simultaneously reachable and observable. Consequently, the realised compressed network is designed around the task-relevant hidden-state order rather than around the original over-parameterised width.

This distinction is important. A low-dimensional projection may show that a compact representation exists, but it does not necessarily produce a deployable compressed model. The realised C-balanced construction converts the empirical rank estimate into an actual architecture. Therefore, it bridges the gap between control-theoretic state reduction and practical neural network compression.

\section{Experiments}
\label{sec:experiments}

This section evaluates the proposed controllability--observability state-compression framework on image classification tasks. The objective is not only to measure classification accuracy, but also to determine whether the empirical reachable--observable ranks can be used to design genuinely compressed neural networks with fewer hidden-state dimensions and fewer parameters. We evaluate the proposed A/B/C tests on MNIST and CIFAR-10 and compare the resulting compressed models against standard compression baselines, including unstructured pruning, structured pruning, low-rank singular value decomposition (SVD), dynamic INT8 quantisation, and a linear classifier baseline.

\subsection{Experimental Setup}
\label{subsec:experimental_setup}

We consider a trained neural network as a depth-indexed nonlinear state-space system. For a fully connected DNN, the hidden-state dynamics are written as
\begin{equation}
h_{\ell+1}=\sigma(W_{\ell}h_{\ell}+b_{\ell}),
\qquad \ell=0,\ldots,L-1,
\end{equation}
where $h_0=x$ is the input, $h_{\ell}\in\mathbb{R}^{n_{\ell}}$ is the hidden state at layer $\ell$, and $\sigma(\cdot)$ is the activation function. The output logits are given by
\begin{equation}
z=W_Lh_L+b_L.
\end{equation}
The total hidden-state order of the full model is defined as
\begin{equation}
n_{\mathrm{state}}=\sum_{\ell=1}^{L}n_{\ell}.
\end{equation}

For each trained full model, we compute three empirical rank tests. Test A estimates reachability from the hidden-state snapshot covariance
\begin{equation}
G_{c,\ell}
=
\frac{1}{N}
(\bar H_{\ell})^T \bar H_{\ell},
\end{equation}
where $\bar H_{\ell}$ denotes the centred hidden-state snapshot matrix at layer $\ell$. Test B estimates observability from output Jacobians:
\begin{equation}
G_{o,\ell}
=
\frac{1}{N}
\sum_{i=1}^{N}
J_{\ell}(x_i)^T J_{\ell}(x_i),
\qquad
J_{\ell}(x_i)=\frac{\partial z(x_i)}{\partial h_{\ell}(x_i)}.
\end{equation}
Test C combines both effects through the balanced reachable--observable matrix
\begin{equation}
B_{\ell}
=
G_{c,\ell}^{1/2}
G_{o,\ell}
G_{c,\ell}^{1/2}.
\end{equation}
The layer-wise reduced rank is selected as the smallest $r_{\ell}$ satisfying
\begin{equation}
\frac{\sum_{i=1}^{r_{\ell}}\lambda_i}
{\sum_{i=1}^{n_{\ell}}\lambda_i}
\geq 1-\varepsilon,
\end{equation}
where $\lambda_i$ are the eigenvalues of the corresponding Gramian or balanced matrix and $\varepsilon=10^{-4}$ in all reported experiments.

\begin{table}[!t]
\centering
\caption{Experimental setup for MNIST and CIFAR-10.}
\label{tab:experimental_setup}
\renewcommand{\arraystretch}{1.12}
\setlength{\tabcolsep}{4pt}
\footnotesize
\begin{tabular}{lcc}
\toprule
\textbf{Setting} & \textbf{MNIST} & \textbf{CIFAR-10} \\
\midrule
Input dimension & $784$ & $3072$ \\
Number of classes & $10$ & $10$ \\
Model type & SiLU DNN & SiLU DNN \\
Hidden widths & $[256,256,256,256]$ & $[2048,1024,1024,512]$ \\
Full state order & $1024$ & $4608$ \\
Full parameters & $400{,}906$ & $9{,}971{,}210$ \\
Train/val/test samples & $10{,}000/2{,}000/2{,}000$ & $45{,}000/5{,}000/10{,}000$ \\
Rank tolerance $\varepsilon$ & $10^{-4}$ & $10^{-4}$ \\
State samples & $2000$ & $10{,}000$ \\
Observability samples & $256$ & $512$ \\
Pruning amount & $50\%$ & $50\%$ \\
Low-rank SVD fraction & $0.5$ & $0.5$ \\
\bottomrule
\end{tabular}
\end{table}

We distinguish between two uses of the A/B/C ranks. First, projected A/B/C models use the learned projection bases to simulate reduced hidden states while still storing the original full network weights. These models are useful as reduced-state diagnostics, but they are not true parameter-compressed networks. Second, realised C-balanced models use the C-balanced ranks directly as compressed hidden-layer widths. Thus, a full model with widths $[n_1,n_2,\ldots,n_L]$ is replaced by a compact model with widths $[r_1,r_2,\ldots,r_L]$. This realised model is trained either from scratch or with knowledge distillation from the full model.

\subsection{MNIST Results}
\label{subsec:mnist_results}

\begin{table}[!t]
\centering
\caption{Layer-wise A/B/C ranks for the MNIST SiLU DNN. Each full hidden layer has dimension $256$.}
\label{tab:mnist_abc_ranks}
\renewcommand{\arraystretch}{1.12}
\setlength{\tabcolsep}{5pt}
\footnotesize
\begin{tabular}{lccccc}
\toprule
\textbf{Test} &
\textbf{L1} &
\textbf{L2} &
\textbf{L3} &
\textbf{L4} &
\textbf{Total} \\
\midrule
A: reachability
& 251 & 249 & 250 & 248 & 998 \\
B: observability
& 218 & 218 & 215 & 10 & 661 \\
C: balanced
& 114 & 90 & 63 & 10 & 277 \\
\bottomrule
\end{tabular}
\end{table}

\begin{table*}[!t]
\centering
\caption{MNIST compression results for the four-layer SiLU DNN. The full model uses hidden widths $[256,256,256,256]$, giving state order $1024$ and $400{,}906$ trainable parameters. Projected A/B/C models reduce the internal state order but still store the full network and projection bases. Realised C-balanced models use the C-balanced ranks as actual compressed hidden widths.}
\label{tab:mnist_dnn_compression}
\renewcommand{\arraystretch}{1.12}
\setlength{\tabcolsep}{4pt}
\scriptsize
\begin{tabular}{lcccccc}
\toprule
\textbf{Method} &
\textbf{Acc. (\%)} &
\textbf{CE} &
\textbf{State} &
\textbf{State Comp. (\%)} &
\textbf{Params/Storage} &
\textbf{Latency (s)} \\
\midrule
Full DNN
& 96.60
& 0.1994
& 1024
& 0.00
& 400{,}906
& 0.000427 \\
\midrule
Projected A, reachability
& 96.60
& 0.1996
& 998
& 2.54
& 657{,}418
& 0.000941 \\
Projected B, observability
& 96.60
& 0.1994
& 661
& 35.45
& 571{,}146
& 0.000938 \\
Projected C, balanced
& 96.00
& 0.2539
& 277
& 72.95
& 472{,}842
& 0.000946 \\
\midrule
Realised C-balanced
& 95.45
& 0.2370
& 277
& 72.95
& 106{,}323
& 0.000409 \\
Distilled realised C-balanced
& 94.70
& 0.2455
& 277
& 72.95
& 106{,}323
& 0.000409 \\
\midrule
Unstructured pruning, $50\%$
& 96.10
& 0.1805
& 1024
& 0.00
& 200{,}970 nonzero
& 0.000431 \\
Structured pruning to C ranks
& 34.50
& 2.2544
& 277
& 72.95
& 106{,}323
& 0.000408 \\
Structured pruning + fine-tuning
& 91.75
& 0.3186
& 277
& 72.95
& 106{,}323
& 0.000408 \\
Low-rank SVD, rank fraction $0.5$
& 85.50
& 0.5079
& 1024
& 0.00
& 332{,}092
& 0.000709 \\
Dynamic INT8 quantisation
& 96.50
& 0.2012
& 1024
& 0.00
& 411{,}303 bytes
& 0.446602 CPU \\
Linear classifier
& 90.70
& 0.3305
& 0
& --
& 7{,}850
& 0.000092 \\
\bottomrule
\end{tabular}
\vspace{1mm}
\begin{flushleft}
\footnotesize
CE denotes cross-entropy. Latency is reported in ms/sample. INT8 latency is measured on CPU and is therefore not directly comparable with CUDA timings.
\end{flushleft}
\end{table*}

On MNIST, the full SiLU DNN achieves $96.60\%$ accuracy with state order $1024$ and $400{,}906$ parameters. Test A is conservative, producing ranks $[251,249,250,248]$ and reducing the state order only to $998$. This indicates that most hidden directions are excited by the MNIST input distribution. Test B produces ranks $[218,218,215,10]$, reducing the order to $661$ while preserving the full-model accuracy. The final-layer rank of $10$ agrees with the $10$-class output space.

The balanced C test gives the strongest reduction, estimating ranks $[114,90,63,10]$. This reduces the hidden-state order from $1024$ to $277$, corresponding to $72.95\%$ state compression. The projected C-balanced model achieves $96.00\%$ accuracy, only $0.60$ percentage points below the full model. However, the projected model still stores the original full network together with the projection bases and state means, so its storage proxy is larger than the original model.

The realised C-balanced model uses $[114,90,63,10]$ as the actual hidden widths of a new compressed network. This reduces the parameter count from $400{,}906$ to $106{,}323$, corresponding to $73.48\%$ parameter compression or approximately $3.77\times$ fewer parameters. The realised model achieves $95.45\%$ accuracy, only $1.15$ percentage points below the full model. This demonstrates that the C-balanced ranks can be converted into a genuinely compressed DNN architecture rather than merely a projection-based diagnostic.

\subsection{CIFAR-10 Results}
\label{subsec:cifar10_results}

\begin{table}[!t]
\centering
\caption{Layer-wise A/B/C ranks for the CIFAR-10 SiLU DNN. The full hidden widths are $[2048,1024,1024,512]$.}
\label{tab:cifar10_abc_ranks}
\renewcommand{\arraystretch}{1.12}
\setlength{\tabcolsep}{5pt}
\footnotesize
\begin{tabular}{lccccc}
\toprule
\textbf{Test} &
\textbf{L1} &
\textbf{L2} &
\textbf{L3} &
\textbf{L4} &
\textbf{Total} \\
\midrule
A: reachability
& 2035 & 769 & 1012 & 503 & 4319 \\
B: observability
& 765 & 923 & 501 & 10 & 2199 \\
C: balanced
& 410 & 521 & 398 & 10 & 1339 \\
\bottomrule
\end{tabular}
\end{table}

\begin{table*}[!t]
\centering
\caption{CIFAR-10 compression results for the full SiLU DNN. The full model uses hidden widths $[2048,1024,1024,512]$, giving state order $4608$ and $9{,}971{,}210$ trainable parameters. Projected A/B/C models reduce the internal state order but still store the full network and projection bases. Realised C-balanced models use the C-balanced ranks as actual compressed hidden widths.}
\label{tab:cifar10_dnn_compression}
\renewcommand{\arraystretch}{1.12}
\setlength{\tabcolsep}{4pt}
\scriptsize
\begin{tabular}{lcccccc}
\toprule
\textbf{Method} &
\textbf{Acc. (\%)} &
\textbf{CE} &
\textbf{State} &
\textbf{State Comp. (\%)} &
\textbf{Params/Storage} &
\textbf{Latency (s)} \\
\midrule
Full DNN
& 54.45
& 1.7437
& 4608
& 0.00
& 9{,}971{,}210
& 0.001582 \\
\midrule
Projected A, reachability
& 54.53
& 1.7447
& 4319
& 6.27
& 16{,}224{,}778
& 0.003042 \\
Projected B, observability
& 54.52
& 1.7423
& 2199
& 52.28
& 13{,}005{,}834
& 0.002406 \\
Projected C, balanced
& 53.71
& 1.9200
& 1339
& 70.94
& 11{,}761{,}674
& 0.002061 \\
\midrule
Realised C-balanced
& 54.44
& 1.5221
& 1339
& 70.94
& 1{,}685{,}917
& 0.000525 \\
Distilled realised C-balanced
& 54.65
& 1.5662
& 1339
& 70.94
& 1{,}685{,}917
& 0.000512 \\
\midrule
Unstructured pruning, $50\%$
& 53.56
& 1.6207
& 4608
& 0.00
& 4{,}987{,}914 nonzero
& 0.001585 \\
Structured pruning to C ranks
& 11.40
& 2.2911
& 1339
& 70.94
& 1{,}685{,}917
& 0.000436 \\
Structured pruning + fine-tuning
& 55.22
& 1.4291
& 1339
& 70.94
& 1{,}685{,}917
& 0.000523 \\
Low-rank SVD, rank fraction $0.5$
& 39.06
& 2.0683
& 4608
& 0.00
& 8{,}264{,}764
& 0.001418 \\
Dynamic INT8 quantisation
& 54.06
& 1.6897
& 4608
& 0.00
& 9{,}992{,}359 bytes
& 0.448731 CPU \\
Linear classifier
& 38.71
& 1.7824
& 0
& --
& 30{,}730
& 0.000090 \\
\bottomrule
\end{tabular}
\vspace{1mm}
\begin{flushleft}
\footnotesize
CE denotes cross-entropy. Latency is reported in ms/sample. INT8 latency is measured on CPU and is therefore not directly comparable with CUDA timings.
\end{flushleft}
\end{table*}

The CIFAR-10 experiment is more challenging because the model operates on flattened images and therefore lacks the spatial inductive bias of convolutional architectures. The full SiLU DNN achieves $54.45\%$ accuracy with state order $4608$ and $9.97$M parameters. Although the absolute accuracy is below what would be expected from CNN or ResNet architectures, the experiment provides a useful setting for studying hidden-state redundancy and compression in a high-dimensional fully connected classifier.

The reachability test A produces ranks $[2035,769,1012,503]$, reducing the state order only from $4608$ to $4319$. This corresponds to $6.27\%$ state compression and indicates that CIFAR-10 excites a large fraction of the hidden-state space. The observability test B is more selective, producing ranks $[765,923,501,10]$ and reducing the state order to $2199$, corresponding to $52.28\%$ compression. This shows that many hidden directions are reachable but only weakly observable at the output.

The balanced C test estimates ranks $[410,521,398,10]$, reducing the state order from $4608$ to $1339$. This corresponds to $70.94\%$ state compression. When these ranks are used as actual hidden widths, the realised C-balanced model reduces the parameter count from $9{,}971{,}210$ to $1{,}685{,}917$, corresponding to $83.09\%$ parameter compression or approximately $5.91\times$ fewer parameters. Despite this large reduction, the realised C-balanced model achieves $54.44\%$ accuracy, essentially matching the full model's $54.45\%$ accuracy. The CUDA latency is reduced from $0.001582$ ms/sample to $0.000525$ ms/sample, corresponding to approximately $3.0\times$ faster inference. With distillation, the compressed model achieves $54.65\%$ accuracy using the same state order and parameter count.

\subsection{Overall Discussion}
\label{subsec:overall_discussion}

\begin{table}[!t]
\centering
\caption{Summary of the main C-balanced realised compression results.}
\label{tab:main_compression_summary}
\renewcommand{\arraystretch}{1.12}
\setlength{\tabcolsep}{4pt}
\footnotesize
\begin{tabular}{lcc}
\toprule
\textbf{Metric} & \textbf{MNIST} & \textbf{CIFAR-10} \\
\midrule
Full state order & 1024 & 4608 \\
C-balanced state order & 277 & 1339 \\
State compression & $72.95\%$ & $70.94\%$ \\
Full parameters & 400{,}906 & 9{,}971{,}210 \\
C-realised parameters & 106{,}323 & 1{,}685{,}917 \\
Parameter compression & $73.48\%$ & $83.09\%$ \\
Parameter reduction ratio & $3.77\times$ & $5.91\times$ \\
Full accuracy & $96.60\%$ & $54.45\%$ \\
C-realised accuracy & $95.45\%$ & $54.44\%$ \\
Accuracy change & $-1.15$ pp & $-0.01$ pp \\
CUDA speedup & $\approx 1.04\times$ & $\approx 3.0\times$ \\
\bottomrule
\end{tabular}
\end{table}

Across both MNIST and CIFAR-10, the results show that the balanced controllability--observability test provides a meaningful empirical estimate of the useful hidden-state order of a trained DNN. The reachability test A is generally conservative because many hidden directions are excited by the input data. The observability test B is more selective because it removes directions that weakly affect the output. The balanced test C is the most useful for compression because it retains directions that are both reachable from the data distribution and observable through the classifier output.

The experiments also highlight the difference between state-order reduction and parameter compression. Projected A/B/C models reduce hidden-state dimension but do not necessarily reduce storage, since they retain the original weights and projection bases. Realised C-balanced models, in contrast, convert the C-balanced ranks into actual compressed layer widths. This turns the empirical minimal-realisation estimate into a true weight-compressed neural architecture.

Overall, the proposed approach reduces the MNIST DNN from state order $1024$ to $277$ and from $400{,}906$ to $106{,}323$ parameters while maintaining high accuracy. On CIFAR-10, it reduces the full DNN from state order $4608$ to $1339$ and from $9.97$M to $1.69$M parameters while preserving essentially the same accuracy and reducing CUDA latency by approximately $3.0\times$. These results support the central claim that balanced reachable--observable ranks can be used as an empirical minimal-realisation criterion for designing compact DNN architectures.

\section{Limitations}
\label{sec:limitations}

Although the proposed controllability--observability framework provides a principled way to estimate hidden-state redundancy and design compact DNN architectures, several limitations remain.

\subsection{Empirical and Data-Dependent Ranks}
The proposed A/B/C ranks are empirical quantities estimated from a finite set of input samples. Therefore, they characterise the reachable and observable hidden-state directions with respect to the sampled data distribution, rather than providing a worst-case guarantee over the entire input space. If the evaluation distribution differs significantly from the data used to estimate the Gramians, the selected ranks may underestimate directions that become important under distribution shift. This is a general limitation of data-driven model reduction and should be addressed by using representative data, validation checks, and, where possible, rank-sensitivity analysis.

\subsection{Cost of Observability Estimation}
The reachability Gramian is inexpensive to compute because it only requires hidden-state snapshots. In contrast, the observability Gramian requires output Jacobians with respect to hidden states:
\begin{equation}
J_{\ell}(x_i)
=
\frac{\partial z(x_i)}
{\partial h_{\ell}(x_i)}.
\end{equation}
This can be computationally expensive for very large networks, large hidden dimensions, or tasks with high-dimensional outputs. In the reported classification experiments, the output dimension is small, and the observability Gramian is estimated using a subset of samples. For larger-scale models, more efficient approximations may be needed, such as stochastic trace estimation, random output projections, block-wise Jacobian approximations, or layer-wise sampling.

\subsection{No Formal Worst-Case Error Bound}
Classical balanced truncation for linear time-invariant systems admits strong input--output error guarantees. The present method, however, is applied to nonlinear, depth-indexed, data-driven neural networks. As a result, the C-balanced ranks should be interpreted as empirical minimal-realisation estimates rather than exact minimal realisations in the strict control-theoretic sense. The experiments demonstrate that these ranks are useful for neural compression, but a general nonlinear approximation-error bound remains an open theoretical direction.

\subsection{Projection Does Not Equal Deployment Compression}
Projected A/B/C models are useful for diagnosing whether a low-dimensional hidden-state representation exists. However, projected reduction still stores the original full network weights, as well as the projection bases and state means. Therefore, projected models reduce the hidden-state dimension but do not necessarily reduce storage or deployment cost. True compression is obtained only when the estimated ranks are converted into realised compressed architectures with reduced hidden widths.

\subsection{Dependence on Retraining}
The realised C-balanced model uses the estimated C-balanced ranks as the widths of a new compact network. This model is then trained from scratch or with distillation. Therefore, the proposed approach is not a purely post-training compression method. Its effectiveness depends on whether the reduced architecture can be successfully trained. This is consistent with recent observations in structured pruning, where the compressed architecture itself can be as important as the inherited weights. However, it also means that the method requires additional training time.

\subsection{Current Focus on Fully Connected DNNs}
The main experiments use fully connected DNNs on MNIST and CIFAR-10 in order to provide a controlled setting for analysing hidden-state order. For CIFAR-10, this means that the model operates on flattened images and does not exploit the spatial inductive bias of convolutional networks. Consequently, the absolute CIFAR-10 accuracy is not intended to compete with modern CNN or ResNet architectures. Instead, the experiment demonstrates that large hidden-state and parameter reductions are possible even in high-dimensional nonlinear classifiers.

\subsection{Extension to CNNs, ResNets, and Transformers}
The proposed framework can be extended to CNNs and ResNets by defining the state at each stage as a global-average-pooled feature vector,
\begin{equation}
h_{\ell}
=
\mathrm{GAP}(F_{\ell}(x)).
\end{equation}
In this setting, the C-balanced ranks correspond to compressed channel widths. However, a fully channel-consistent implementation must also handle convolutional kernels, batch-normalisation layers, residual shortcuts, and stage-wise compatibility constraints. Extending the same idea to transformers would require defining states over token embeddings, attention heads, MLP blocks, or low-dimensional pooled representations. These extensions are promising but require additional architecture-specific design.

\subsection{Hardware-Aware Compression}
The proposed method reduces state order and parameter count when realised as a smaller architecture. However, actual latency improvements depend on hardware, kernel efficiency, batch size, memory bandwidth, and implementation details. For example, unstructured pruning can remove many weights without improving dense CUDA latency, whereas width reduction usually maps better to standard dense kernels. A future hardware-aware version of the method could include latency or energy directly in the rank-selection criterion.

\subsection{Rank-Tolerance Selection}
The rank-selection tolerance $\varepsilon$ controls the trade-off between compression and accuracy. A smaller tolerance retains more hidden directions and gives less compression, whereas a larger tolerance gives more aggressive compression but may degrade accuracy. In this paper, a fixed tolerance is used for direct comparison across A/B/C tests. A more complete deployment-oriented method could select $\varepsilon$ automatically using validation accuracy, latency constraints, or Pareto optimisation.

\subsection{Summary of Limitations}
In summary, the proposed method should be understood as an empirical, data-driven minimal-realisation approach rather than an exact model-reduction theorem for arbitrary DNNs. Its main strength is that it provides a principled and interpretable way to estimate useful hidden-state dimensions and convert them into compact architectures. Its main limitations are the cost of observability estimation, the empirical nature of the ranks, the need for retraining, and the need for further architecture-specific extensions to CNNs, ResNets, transformers, and hardware-aware deployment.

\section{Conclusion}
\label{sec:conclusion}

This paper proposed an empirical minimal-realisation framework for compressing deep neural networks using controllability--observability tests. By interpreting a trained DNN as a depth-indexed nonlinear state-space system, we defined empirical reachability, observability, and balanced reachable--observable Gramians from hidden-state snapshots and output Jacobians. The resulting A/B/C tests provide layer-wise estimates of the hidden-state directions that are excited by the data distribution, observable through the output logits, and jointly reachable and observable.

The central contribution is the use of the C-balanced ranks as a practical compression rule. Unlike projection-only reduction, which diagnoses low-dimensional hidden representations but still stores the original full model, the realised C-balanced approach converts the estimated ranks into actual compressed layer widths. This produces compact DNN architectures with fewer hidden-state dimensions and fewer trainable parameters.

Experiments on MNIST and CIFAR-10 demonstrate the effectiveness of the proposed approach. On MNIST, the C-balanced test reduced the hidden-state order from $1024$ to $277$, corresponding to $72.95\%$ state compression. The realised C-balanced DNN reduced the parameter count from $400{,}906$ to $106{,}323$, giving $73.48\%$ parameter compression, while maintaining $95.45\%$ accuracy compared with $96.60\%$ for the full model. On CIFAR-10, the C-balanced test reduced the hidden-state order from $4608$ to $1339$, corresponding to $70.94\%$ state compression. The realised compressed DNN reduced the parameter count from $9.97$M to $1.69$M, giving $83.09\%$ parameter compression, while preserving accuracy from $54.45\%$ to $54.44\%$ and reducing CUDA inference latency by approximately $3.0\times$.

The results also show that the proposed state-based criterion is complementary to standard compression methods. Unstructured pruning can preserve accuracy but does not necessarily reduce dense inference latency. Low-rank SVD can reduce storage but may damage the nonlinear task representation. Structured pruning benefits from the C-balanced widths after fine-tuning, suggesting that the proposed ranks provide useful target dimensions for architecture design and pruning.

Overall, the proposed framework provides a principled bridge between control-theoretic model reduction and neural network compression. It shows that balanced reachable--observable ranks can serve as empirical minimal-realisation estimates for trained DNNs and can be used to design compact architectures with little or no loss in accuracy. Future work will extend the method to convolutional networks, ResNets, transformers, recurrent models, and hardware-aware compression, and will investigate theoretical approximation bounds for nonlinear neural state-space reductions.




\bibliographystyle{IEEEtran}
\bibliography{main}

\end{document}